%% file: main.tex
\documentclass[runningheads]{llncs}
\usepackage{amssymb,mathtools}
\usepackage{graphicx}
\usepackage{booktabs}       % professional-quality tables
\usepackage{xcolor}         % colors
\begin{document}

\definecolor{carnelian}{rgb}{0.7, 0.11, 0.11}
\definecolor{nc}{rgb}{0.2, 0.75, 0.3}
\newcommand{\cv}[1]{{\bfseries \color{carnelian} #1 }}
\newcommand{\ag}[1]{{\bfseries \color{blue} #1 }}
\newcommand{\wl}[1]{{\bfseries \color{orange} #1 }}
\newcommand{\nc}[1]{{\color{nc} #1 }}
\title{Predictive maintenance on event logs: Application on an ATM fleet}
\author{Antoine Guillaume$^{1,2}$, Christel Vrain$^1$, Wael Elloumi$^2$}

\authorrunning{A. Guillaume et al.}
  
\institute{LIFO EA 4022, Univ. Orl\'eans \and Worldline}

\maketitle              % typeset the header of the contribution
\begin{abstract}
Predictive maintenance is used in industrial applications to increase machine availability and optimize cost related to unplanned maintenance. In most cases, predictive maintenance applications use output from sensors, recording physical phenomenons such as temperature or vibration which can be directly linked to the degradation process of the machine. However, in some applications, outputs from sensors are not available, and event logs generated by the machine are used instead. We first study the approaches used in the literature to solve predictive maintenance problems and present a new public dataset containing the event logs from 156 machines. After this, we define an evaluation framework for predictive maintenance systems, which takes into account business constraints, and conduct experiments to explore suitable solutions, which can serve as guidelines for future works using this new dataset.

\keywords{Predictive Maintenance \and Time Series \and Change point detection.}
\end{abstract}
\input{Introduction.tex}
\input{Background.tex}
\input{Dataset.tex}

\input{Experiments.tex}
\input{Conclusion.tex}

\bibliographystyle{splncs04}
\bibliography{biblio.bib}

\end{document}

%% file: Introduction.tex
\section{Introduction}
The main goal of predictive maintenance is to increase machine availability by minimizing unplanned maintenance caused by machine failures. Knowing when and why a machine is going to fail offers many benefits, such as a better maintenance planning, stock part management, and other cost optimization related to the maintenance process. Other benefits vary depending on the use case: for example, in a factory it can lead to increase productivity, whereas in a hospital it can reduce the downtime of machines, which may be critical to patient health.

A classical example of predictive maintenance is the case of vibration monitoring \cite{ORHAN2006293}. Consider a rotating machinery in an industrial context, equipped with sensors that can measure the amplitude of vibrations emitted by the rotating piece. In such signals, the degradation process of the machine can often be clearly modeled as linear or exponential. This information can then be used to identify ongoing degradation, and with historic knowledge of the breaking point (i.e. the amplitude where pieces have broken in the past), build a model to estimate the remaining time before failure, based on recent sensor data. More complex systems can then be monitored by including more parameters, like temperature or pressure measurements.

%Through this paper, we consider the case where the machines we want to monitor are not equipped with sensors to capture meaningful features such as vibration, and it is not reasonable or possible to install such sensors, or to acquire the outputs of the sensors. 
%Therefore, we leverage the logs produced by those machines to identify patterns that allow to predict equipment failures. The main difference between sensors and logs based approaches is that log data often represent logical events, rather than the measurement of a physical event. Even if those logical events are triggered by physical events, the information they provide is often discrete and entails an information loss regarding the physical event linked to the degradation process. 

In the following, we study the case of a fleet of automated teller machines (ATM), for which we do not have any sensor data, but rather event logs that offer a high level view of the physical events happening on an ATM. The main challenge of this application, is to identify relevant patterns in series of discrete logical events (i.e event logs), where one event log can be triggered by multiple physical phenomenons. Our contributions can be summarized as follows:

\begin{itemize}
\item A new difficult dataset of log data from a real industrial ATM fleet, publicly available and presented in Section \ref{sec:Dataset}.
\item An empirical study comparing the performance of different change point detection methods, setting baselines for future experiments using this dataset.
\item A metric based on business constraints, to estimate the efficiency of a predictive maintenance system.
\end{itemize}

%% file: Background.tex
\section{Related Works}
The field of predictive maintenance has been very active in the past decade. A recent survey on predictive maintenance \cite{SurvPDM1} shows the huge diversity of solutions used to solve the predictive maintenance problem. Many different approaches exist, with the common point that those approaches are adapted to the case of temporal data, which composes most, if not all, inputs for predictive maintenance problems. We summarize those approaches as follows:

\begin{itemize}
\item \textbf{Data Mining approaches}. Those approaches generally use a range of data mining related techniques, such as temporal rules mining with chronicles mining \cite{Chronicles}, or rules learned from the data as in \cite{eGFC} where the authors introduce an online semi-supervised approach for anomaly detection using logs from data centers. We also find change point detection methods \cite{ChangePoint}, which try to estimate when a machine starts to malfunction.

\item \textbf{Model based approaches}. This category includes the use of techniques such as survival analysis, also known as hazard-rate models \cite{SurvivalPdM}, autoregressive models \cite{PdMARMA} or Markov models. The goal is to model the behavior of the monitored machine to identify deviations from normal behavior or estimate a path-to-failure.

\item \textbf{Machine learning approaches}.
Some design features by hand, typically with sliding window statistics, in order to remove the temporal component and to apply classic algorithms \cite{PdMATM}. The use of time series transformation techniques such as ROCKET\cite{ROCKET} or MrSEQL\cite{MrSEQL} is also possible to extract features which can be used to spot anomalies. Lastly, Deep Learning methods \cite{TSCDeepRev} such as Convolutional or Recurrent Network can be used to deal with the time component. Auto-Encoders can also be considered to perform anomaly detection \cite{AutoEncoders}.
\end{itemize}

The choice of the right approach and problem formulation is dependent on multiple factors such as the knowledge of the degradation process of the machines (e.g linear, exponential, random, unknown), the properties of the time series data, business constraints, and the quantity and quality of labeled data.

\subsection{Problem formulation}
\label{sec:PbForm}
While formulating a predictive maintenance problem, it is common to define a set of time intervals to include business constraints. It does not seem to exist a consensus for the definitions of those intervals in the literature, even if the logic behind them is mostly the same. We choose to base ourselves on the definitions of \cite{Sipos2014} as they best fit our vision of the problem formulation. 

Let us consider a time series $X: \{(t_0, x_0), ..., (t_n, x_n)\}$ of length $n$ representing the data of a machine, with $t_i$ a timestamp and $x_i$ an event. Supposing that no failure has occurred between $t_0$ and $t_n$, we want to know if a failure will happen in the near future after $t_n$. To better frame this statement and to take into account business constraints we need to define three time intervals, which can be used for data preprocessing, labeling or performance metrics:

\begin{itemize}
\item \textbf{Responsive Duration} $(rd)$: a time interval reflecting the real-life time needed to perform maintenance. It can be measured as the mean time from the moment an alert is raised to the end of the maintenance process on-site.

\item \textbf{Predictive Padding} $(pp)$: a time interval reflecting how much in advance it is acceptable to predict a failure. %\cv{Je ne suis pas certaine que cette définition soit équivalente à la période dans laquelle existe la signature de panne}\ag{Ce n'est plus l'objectif, mais effectivement certain passage prete a confusion, je corrige}

\item \textbf{Infected Interval} $(ii)$: a time interval which reflects the time needed to return to normal behavior after a failure.
\end{itemize}

In most approaches, the notion of infected interval aims at removing noisy data or data that could be wrongly identified as an abnormal behavior. The use of the other intervals will vary depending on the approach. For example, in a regression context, when predicting the remaining useful life of a machine, it can be directly included as threshold below which an alert will be raised. Whereas for unsupervised approach, it may only be used as a mean to evaluate the efficiency of the system.

%\cv{Dire comme c'est traité dans notre cas}\ag{Ajout petite phrase}
%For example in a binary classification task, if $X$ is known to fail at $t_n$, we could consider data in the interval $[t_n - (pp + rd), t_n - rd[$ as labeled positive, and data before $t_n - (pp + rd)$ as negative. The goal being to make the classification model learn to recognize future failure based on recent data from a machine. One hypothesis made by this formulation is that failure signature only happen in $[t_n - (pp + rd), t_n - rd[$. In the case of regression, we would raise an alert when the predicted remaining life duration will be below $pp+rd$. 

%% file: Dataset.tex
\section{Dataset}
\label{sec:Dataset}
The dataset, along with the code for the experiments is available in the associated GitHub repository \footnote{https://github.com/baraline/ATMpaper2021}, before we give details about the ATM dataset, we define some domain specific terms:

\begin{itemize}
%\item \textbf{Cash-In-Transit (CIT)}: Security companies, which are in charge of moving cash and emptying or refilling ATMs when needed.
\item \textbf{ATM module}: An ATM is an ensemble of connected modules, each composed of several components directly interacting with each other. For example, the distribution module is composed of storage boxes for the bank notes, linked to an ensemble of rolling or suction cup mechanisms to route bank notes to the exit slot.
\item \textbf{ATM life cycles}: Given a particular module of an ATM, we call a life cycle the time interval starting after a maintenance of the module, and ending at the next failure of this same module.
\item \textbf{K7} : Internal abbreviation designating the banknote storage boxes.
%\item \textbf{Withdrawal}: This term simply denote the action, initiated by a user, of withdrawing cash from an ATM. While an event code is linked to this action due to the change in the amount of cash in the safe, no transaction or user data is present in the dataset.
\end{itemize}

\subsection{Dataset description}
\label{ATMdata}
\begin{figure}[h]
  \includegraphics[width=0.95\textwidth]{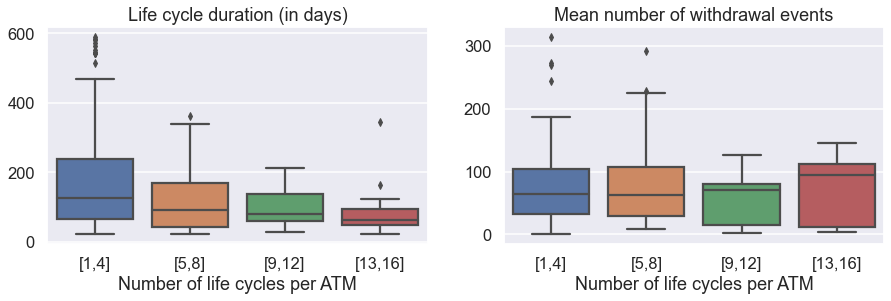}
  \centering
  \caption{Statistics on (a) the duration of life cycles in days and (b) the mean daily number of withdrawal events, grouped by the number of life cycles generated per ATM.}
  \label{fig:datastats}
\end{figure}

The ATM dataset is composed of 292 life cycles of the distribution module, gathered from 156 ATMs located in France over a period of two years, representing more than 12 millions logs. Due to the nature of the application, life cycles have different lengths, with some ATMs containing more life cycles than others. Figure \ref{fig:datastats} gives some statistics about the duration of life cycles and the mean daily number of withdrawal events, contextualized by the number of life cycles an ATM have experienced. 

A life cycle is represented by a univariate time series of uneven frequency, containing event codes as a categorical feature, with 284 unique values. Each value represents an event code, which is linked to a particular logical event on the ATM. As each ATM has its own set of hardware and software, some differences exist in the distribution of event codes between ATMs. This can even occur between the life cycles of an ATM, as the components are subject to change due to maintenance processes or updates.

Data located in time intervals where an ATM was in a failure state, due to problems on other modules, was removed as it is noisy and irrelevant to the task. Note that the notion of infected interval presented in Section \ref{sec:PbForm} was applied with a value of one day independently of the module that caused the failure. %\cv{Cela suppose que la panne est réparée en 1 jour ?}\ag{Non car c'est "after a failure", donc 1+ le temps de panne}

When expressing the state of a hardware component, the event codes generally follow a severity level as OK, Warning or Error, but some do not have a Warning event. For example a code $6000$ indicates an OK event on the distribution module, which can (not exclusively) happen during restart processes or when the ATM recovers from a problem. The code $6001$ indicates an Error event and code $6002$ a Warning event.

An error recovery might for example happen when a banknote is damaged when manipulated by a mechanism. Normally in this case, the ATM is able to reject this banknote into a designated box and avoid further problems. One of the main challenge of the data is that an event code can be triggered by multiple physical events, which make the interpretation of the data difficult.
%\ag{Ajout example}\cv{Est-ce que cela suppose que l'ATM a parfois des erreurs, par exemple de logiciel qu'il peut lui-même corriger} \cv{Peut-être insister ici sur le comportement d'un ATM et la difficulté à interpréter les codes.}

\subsection{Preprocessing} 
For this paper, we choose to focus ourselves on numerical applications rather than approaches such as chronicles mining \cite{Chronicles} than can handle categorical data. To convert the categorical event codes into a numeric form, we create a new feature for each event code representing a boolean indicator of occurrence. Then, considering a resampling frequency $F$, we resample the data, summing each boolean indicator to obtain a number of occurrences of each category in time windows of size $F$.

\begin{figure}[h]
  \includegraphics[width=0.95\textwidth]{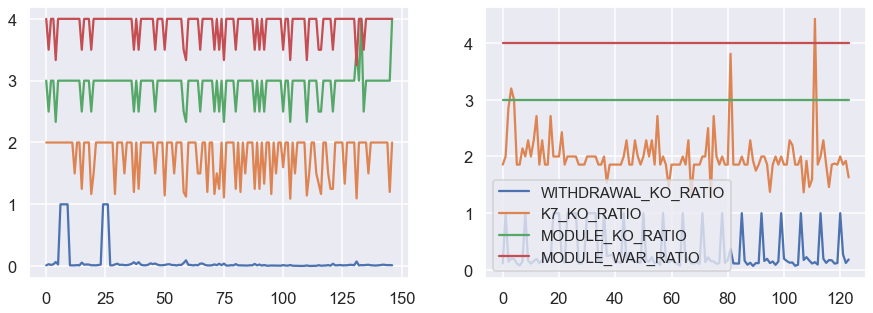}
  \centering
  \caption{Two life cycle time series of the preprocessed dataset with $F=24$ hours, with each feature offseted by 1 for easier visualization.}
  \label{fig:ExPreprocess}
\end{figure}

Then, we make the hypothesis that events happening on a module do not influence the event codes generated by the others, and as our objective is to predict failures of the distribution module, we discard all event codes that are not related to this module. 
%\cv{Cela suppose que les modules sont indépendants et que les codes de warning de l'un ne peuvent influencer l'autre} \ag{Modification}
This reduce the data to 15 event codes, which can be grouped to count the number of OK and Error events for withdrawals, for each of the 5 banknote storage boxes and the distribution module, plus the ratio between OK and Warning events for the distribution module. Figure \ref{fig:ExPreprocess} gives a visualization of the 4 features (with an offset for visualization) computed on the resulting time series with $F=24$ hours. The features represent the ratio of the number of Warning or Error events to the number of OK events, the objective being to detect peaks of Error or Warning events the ATM could not recover from automatically.%\cv{Pas clair}\ag{OK ?}

%% file: Experiments.tex
\section{Experiments}
%\ag{Point de vigilance
%\begin{itemize}
%\item Détails sur le protocole, pourquoi découpage en N jours.
%\item Notion TP,FP,TN,FN avec pp rd 
%\item Comment on score avec le découpage en fenetre
%\item Difference change point detection / anomaly detection.
%\item Sensibilité au seuil
%\item conclusion : opportunité d'apprendre le seuil, ou de méthodologie plus complexe( seuil adaptatif etc...)
%\item faire diff entre E socre par cycle et E score "gobale"
%\end{itemize}
%}
\subsection{Choice of problem formulation}
Multiple constraints must be taken into account to choose the right approach: 
\begin{itemize}
\item The diversity of behaviors, caused by hardware and software differences, as well as differences in frequencies of user interaction and the machine localization.
\item The weak relation between event codes, representing abstract logical events and the physical occurrence of failures.
\item The difference in time series lengths and event patterns.
%\item Industrial and business constraints, where the use of an easily maintainable and interpretable model are preferred for project industrialization.
\end{itemize}

Previous experiments on this problem used a clustering approach to group ATMs with similar behavior and characteristics. The goal was to build a classification model using the life cycles of ATMs within each cluster to facilitate the classification problem. Input data was extracted using sliding windows of two weeks to obtain time series of equal length, which were then labeled as either positive or negative (i.e with a soon to happen failure or not) using the notion of predictive padding. Those approaches were discarded due to the high number of early alerts, caused by the fact that anomalies were present in windows labeled as negative. For the same reason, anomaly detection methods such as \cite{MP}, raised a high number of early alerts. 

To avoid the shortcomings of these approaches, we propose to use change point detection methods. Those methods try to find points in the time series where a change in behavior occurs, which in the context of predictive maintenance, correspond to a degradation of the machine that change the distribution of the data. Additionally, it is an unsupervised approach with fast algorithms and interpretable results, which consider each life cycle independently, thus eliminating the difficult problem of finding common characteristics or patterns among ATMs with a variety of behaviors.

\subsection{Experimental protocol}
As we chose to use unsupervised change point detection methods, each life cycle can be considered independently. Nevertheless, we need to account for the fact that, in our industrial application, the data of a life cycle is a stream updated daily until failure occurs. 

To simulate this, we progressively increase the length of each life cycle by 7 days and try to detect a change point at every step. This process will stop at the first alert raised: this choice is made because in an industrial environment, once an alert is raised, a maintenance will be triggered which make further alerts for the current life cycle worthless.

More formally, consider a preprocessed life cycle $X=\{x_1, \ldots, x_n\}$ with $x1$ a vector containing the features at the first timestamp. Following our protocol, we apply the change point detection methods on $x_1, \ldots, x_T$ (with $T=7$), then on $x_1, \ldots, x_{2T}$, until all the life cycles have been covered or an alert is raised in one of those windows. If the first and only alert is raised during $[n-(pp+rd), n-rd[$, then $X$ will be considered as a True Positive. If any alert is raised before $n-(pp+rd)$ or that the first alert is between $[n-rd, n]$, $X$ will be a False Positive. If no alert was raised in all the 7 day increasing windows, then $X$ is considered a False Negative. 

Note that this formulation would only allow for True Negatives if a life cycle was not complete (i.e yet without failure) and that no alert was raised in all windows. As all life cycles in our dataset have lead to a failure, we will never have True Negative. This protocol is extremely penalizing towards errors: the system must detect a single alert before failure, and this alert has to be in the predictive padding to be considered successful. 

Given an input time series, the goal of change point detection methods is to divide it into sequences or segments, such as each segment isolates a different property of the input. This is often quantified as a change in the data distribution, which translate into the idea of a change in the process generating the data. We choose to consider the 5 following change point detection methods for our experimental protocol:
\begin{itemize}
\item Fast Low-cost Unipotent Semantic Segmentation (\textbf{FLUSS}) \cite{FLUSS}, this algorithm uses the output of  Matrix Profile (MP) \cite{MP} to compute possible segmentation based on the location of similar subsequences across the time series. We impose a threshold on the scoring vector it generates to raise an alert, instead of considering the global minimum.
\item Kernel change point detection (\textbf{KCPD}) \cite{KCPD}. Based on a kernel function, the input time series is mapped to a Hilbert space, and segmentation is performed by finding the partitioning that minimizes a cost function into this space.
\item Linearly penalized segmentation (\textbf{Pelt}) \cite{Pelt}. This algorithm enumerates all possible partitions and finds the segmentation that minimizes a cost function. It uses pruning rules that allow this enumeration to be made in linear time.
\item Binary segmentation (\textbf{Binseg}) \cite{Binseg}. This method uses a greedy procedure, first finding a change point that minimizes a cost function on the whole time series, and then repeats this procedure on the two partitions created. 
\item Bottom-up segmentation (\textbf{BottomUp}) \cite{BottomUp}. Doing the opposite of Binseg, this method starts by identifying a lot of change points, and successively deletes the ones that are less significant.
\end{itemize}

For each method, we test a wide range of parameters, notably the cost functions and the penalty or threshold, which are the main components influencing the results. The complete list of parameter combinations used ($\sim 300$ per model) and results for each cycle is available as supplementary material in the online repository.

\subsection{Metrics for predictive maintenance}
In the context of predictive maintenance, evaluating the performance of a system requires some adjustments. Consider the timestamps of a life cycle $X = \{t_0, ..., t_n\}$, and a model $M$ such as $M(X) = \hat{y}$, with $\hat{y} \in [t_0, t_n]$ a vector giving the locations of the alerts raised by $M$.
If $\hat{y}$ contains multiple alerts, considering all of them in the evaluation metric would introduce a bias compared to the performance of the system when applied in the field. In a real system, any alert will trigger a maintenance that will modify the behavior of the machine, which is why we discard all alerts after the first for evaluating the system performance.

\begin{figure}[h]
  \includegraphics[width=0.95\textwidth]{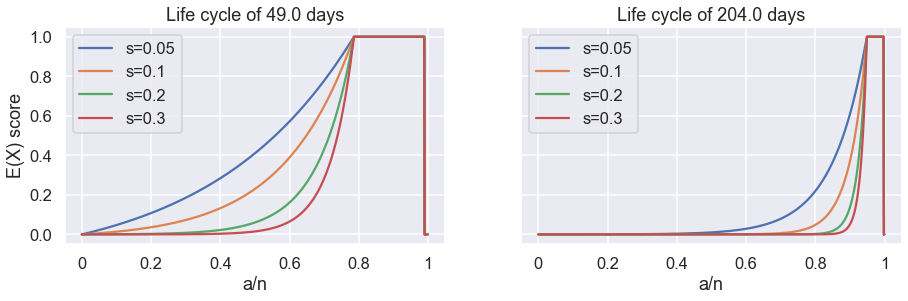}
  \centering
  \caption{Result of $E_s(X)$ for two life cycles of different length and values of $s$, using $pp=10$ and $rd=1$}
  \label{fig:Escore}
\end{figure}

We propose a metric that takes into account the business constraints imposed by the definition of predictive padding ($pp$) and reactive duration ($rd$) in \ref{sec:PbForm}, the difference of length of the life cycles, and can be customized depending on the tolerance to alerts raised before the predictive padding interval. For one life cycle $X$, it takes as input the location of the alert $a$, the length of the life cycle $n$, the predictive padding ($pp$) and reactive duration ($rd$) and a sensibility parameter $s$ to control the tolerance to early alerts. Equation \ref{eq:E} gives the formal definition of this metric, and Figure \ref{fig:Escore} shows a visual representation of the influence of $s$:

\begin{equation}
\label{eq:E}
  E_s(X) =
  \begin{cases}
    0 & \text{if $a \geq n-rd$} \\
    1 & \text{if $n-rd > a \geq n-(rd+pp)$} \\
    \displaystyle{\frac{\exp(a/n)^{n \times s} - 1}{\exp(n-(rd+pp))^{n\times s} - 1}} & \text{otherwise} \\
  \end{cases}
\end{equation}

To evaluate a dataset using this metric, we take the mean $E_s(X)$ score for all samples, giving a percentage that can be interpreted as the degree of optimality of the system given the specified parameters. Our methodology could be compared to the one of \cite{NAB}, which favors systems that identify anomalies in an acceptable range, and only consider the earliest detection to score an anomaly. 

We also use the definitions of \ref{sec:PbForm} to define a binary confusion matrix, with false positive being cycles in which an alert was raised before the predictive padding or in responsive duration, false negative cycles in which no alerts was raised before failure, and true positive where the only alert was raised in the predictive padding interval. This allows us to use metrics such as Precision and Recall to evaluate the system with respect to the business constraints. Again, note that true negative would only happen if we had life cycles in which failure did not yet occur, but we do not have such cycles in the dataset.

\subsection{Experimental results}
\begin{figure}[h]
  \includegraphics[width=0.95\textwidth]{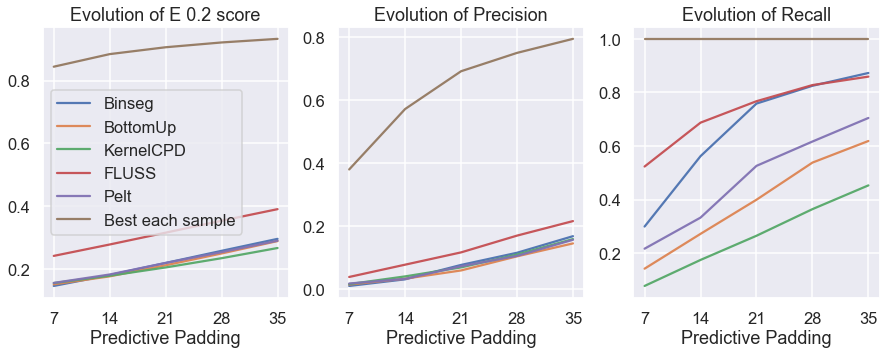}
  \centering
  \caption{Evolution of metrics with different values of the predictive padding.}
  \label{fig:Rank}
\end{figure}

We set $rd=1$ and $s=0.2$ as metric parameters, as they best fit the business use case behind this dataset, we ideally want $pp=14$ in our application, but show the results with different values of this parameters. Figure \ref{fig:Rank} gives the results of each algorithm for the best average parameter combination, and for the best method and parameter combination at each sample. Additional analysis, notably on parameter sensitivity, are provided as supplementary material in the online repository \footnote{https://github.com/baraline/ATMpaper2021}.

As our protocol defines a hard evaluation context, no methods has satisfying results, especially for Precision due to the very strict conditions for a life cycle to be considered as a True Positive. Anyway, it is unrealistic to expect a single method or a single parameter set to cover all samples given the diversity of behaviors between the ATMs.
Z-normalizing the time series at each step of the validation process also slightly increases the performance of all methods, but most notably, it makes FLUSS dominant compared to other others, with it being beaten in only $1\%$ of all life cycles rather than in $15\%$ without normalization.

If we suppose that we had an informed choice of the best method with its set of parameters, that is if at each sample we consider the best method and parameter combination based on the $E_{0.2}$ score, and that we average the result, we obtain nearly perfect score, as show by the "Best each sample" in Figure \ref{fig:Rank}. Of course in reality, this optimal configuration for each life cycle cannot be known. 

Nevertheless, 37\% of ATMs with more than one life cycle share the interesting property that the best model stays the same for all its life cycles. This number goes up to 60\% if we allow one change of model between all life cycles for ATMs with more than two cycles. This shows that we could improve the results by adaptively tuning the choice of parameters and algorithm similarly to \cite{ParamEstim}. This could for example be done using the characteristics of past life cycles of an ATM, or comparing recent data, but this fall out of the scope of this paper.

\begin{figure}[h]
  \includegraphics[width=0.95\textwidth]{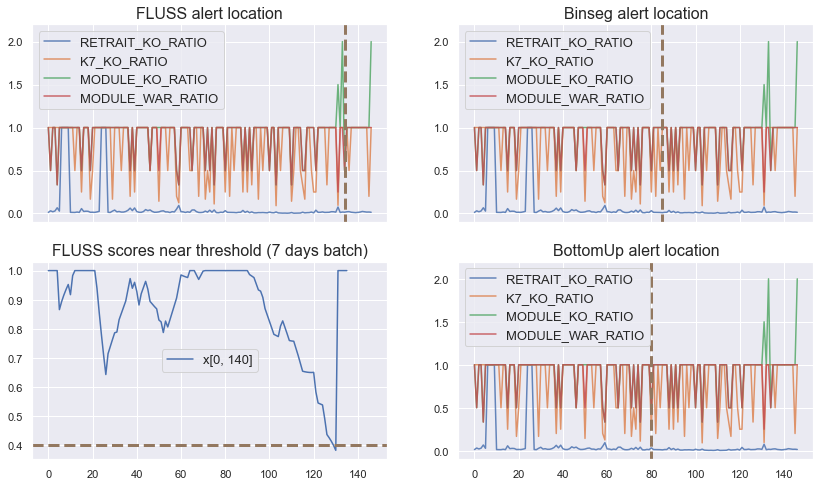}
  \centering
  \caption{Alert location (dashed line) in life cycle n°1 for FLUSS and the other two best methods, with the FLUSS scores close (+0.025) to the threshold for the 7 days batch protocol}
  \label{fig:ex1}
\end{figure}

Looking more closely at the result of all life cycles we can distinguish some interesting patterns. Some life cycles show clear anomalies, only occurring at the end of the life cycle, which makes the identification of the events that lead to a failure straightforward. For example in \ref{fig:ex1}, we clearly see an unprecedented rise in the ratio of Error events (KO in the figure legend) of the distribution module. 

\begin{figure}[h!]
  \includegraphics[width=0.95\textwidth]{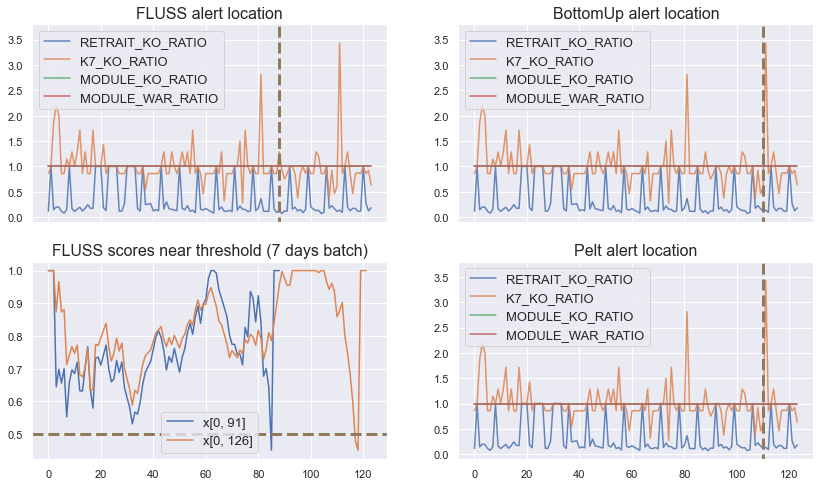}
  \centering
  \caption{Alert location (dashed line) in life cycle n°56 for FLUSS and the other two best methods, with the FLUSS scores close (+0.025) to the threshold for the 7 days batch protocol}
  \label{fig:ex2}
\end{figure}

In other cycles, as shown in Figure \ref{fig:ex2}, we can see that an increase in Error events of the banknote storage (K7 in legend) is likely responsible for the failure. But if we consider this cycle in a streaming context, it may be difficulty to know which peak of Error event is to be considered worth raising an alert. Other life cycles do not exhibit any clear patterns before the end of the life cycle. A deeper inspection is necessary to determine if some kinds of, possibly multivariate, temporal pattern exists that would justify the failure. Another possibility being that external factors, that could not be recorded through the event logs, have caused the failure. 

%% file: Conclusion.tex
\section{Conclusion}
We presented a new publicly available dataset containing 292 life cycles of the distribution module from 156 ATMs, which offer interesting challenges and opportunities for multiple domains. We presented different ways to treat the predictive maintenance problem, and conducted experiments on this new dataset by using change point detection methods, which allowed us to bypass some of the difficulties imposed by the variability of behavior between life cycles.
In future work, we wish to study the possibility to tune the choice of algorithms and parameters adaptively based on the past data and characteristics of each life cycle, either through a learning procedure or the use of heuristics using domain knowledge. The use of fuzzy logic system could also be very adapted to this problem, as it could lessen the effect of irrelevant anomalies that naturally occur in the data.